\documentclass[10pt,twocolumn,letterpaper]{article}

\newcommand{\degree}{$^{\circ}$}

\newcommand{\sphtr}{SPHTR}

\newcommand{\smnist}{SPH-MNIST}
\newcommand{\scifar}{SPH-CIFAR}

\newcommand{\comment}[1]{}
\usepackage[noblocks]{authblk}
\usepackage{cvpr}              

\usepackage{graphicx}
\usepackage{amsmath}
\usepackage{amssymb}
\usepackage{comment}
\usepackage{booktabs}

\usepackage{amsthm}
\theoremstyle{definition}
\newtheorem{definition}{Definition}

\usepackage[T1]{fontenc}
\usepackage{CJKutf8}
\usepackage[english]{babel}
\usepackage{multirow}

\usepackage[pagebackref,breaklinks,colorlinks]{hyperref}

\usepackage[capitalize]{cleveref}
\crefname{section}{Sec.}{Secs.}
\Crefname{section}{Section}{Sections}
\Crefname{table}{Table}{Tables}
\crefname{table}{Tab.}{Tabs.}

\usepackage{array}
\makeatletter
\newcommand{\thickhline}{%
    \noalign {\ifnum 0=`}\fi \hrule height 1pt
    \futurelet \reserved@a \@xhline
}
\newcolumntype{"}{@{\hskip\tabcolsep\vrule width 1pt\hskip\tabcolsep}}
\makeatother

\begin{document}
\begin{CJK}{UTF8}{mj}
\title{Spherical Transformer}

\author{Sungmin Cho$^{1*}$\qquad Raehyuk Jung$^{2*}$\qquad Junseok Kwon$^{3}$ \\
$^{1,3}$School of Computer Science and Engineering, Chung-Ang University, Seoul, Korea\\
$^{2}$Twelve Labs, Seoul, Korea\\
{\tt\small $^{1}$csm8167@naver.com, $^{2}$ray@twelvelabs.io, $^{3}$jskwon@cau.ac.kr}
}

\maketitle
\def\thefootnote{*}\footnotetext{These authors contributed equally to this work}\def\thefootnote{\arabic{footnote}}

\begin{abstract}
Using convolutional neural networks for 360\degree~images can induce sub-optimal performance due to distortions entailed by a planar projection. The distortion gets deteriorated when a rotation is applied to the 360\degree ~image. Thus, many researches based on convolutions attempt to reduce the distortions to learn accurate representation.
In contrast, we leverage the transformer architecture to solve image classification problems for 360\degree~images. Using the proposed transformer for 360\degree~images has two advantages. First, our method does not require the erroneous planar projection process by sampling pixels from the sphere surface. Second, our sampling method based on regular polyhedrons makes low rotation equivariance errors, because specific rotations can be reduced to permutations of faces.
In experiments, we validate our network on two aspects, as follows. First, we show that using a transformer with highly uniform sampling methods can help reduce the distortion. Second, we demonstrate that the transformer architecture can achieve rotation equivariance on specific rotations. 
We compare our method to other state-of-the-art algorithms using the SPH-MNIST, SPH-CIFAR, and SUN360 datasets and show that our method is competitive with other methods. 

\end{abstract}

\section{Introduction}
\label{sec:intro}

\begin{figure}[h]
    \includegraphics[width=\linewidth]{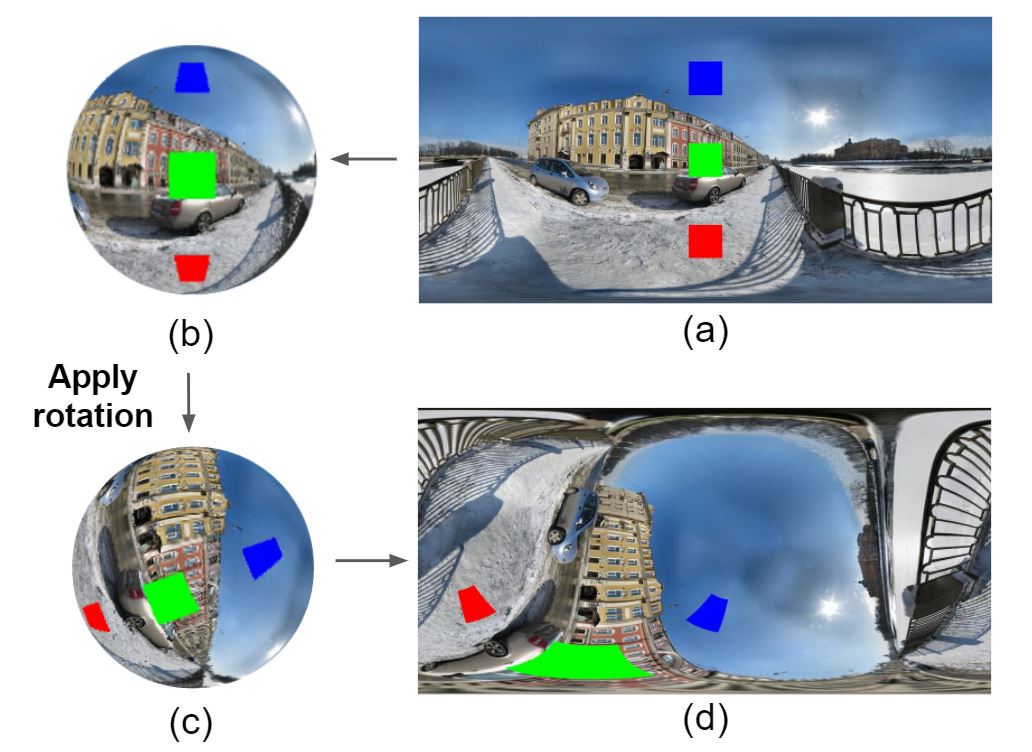}
    \caption{
    \textbf{Examples of projection and rotation distortions.} Red, green, and blue squares represent local patterns.
    (a) The local patterns are visualized in the ERP image. 
    (b) \textit{Projection distortion}: if we visualize local patterns with spherical representation, the patterns have different shapes according to their latitudes, although they have a same shape in (a). 
    (c) A 3D rotation is applied to (b) resulting in (c). 
    (d) \textit{Rotation distortion}: if we visualize (c) in the ERP image, we can witness severe deformation of the local patterns due to projection and rotation distortions.}
    \label{fig:PD_RD}
\end{figure}

Spherical signal is a function that can map points on a sphere to real valued vectors (\ie $f:S^2\rightarrow\mathbb{R}^d$), while they have been widely used for astronomy~\cite{healpix}, robotics~\cite{bazin_omni_robot}, and computer vision~\cite{omni_detection}. In computer vision and robotics, spherical signals are typically represented as 360\degree~images. For example, Google map provides 360\degree~images for road map functionality with variable viewing direction. 

To learn representations of spherical signals using convolutional neural networks (CNNs), we need to transform the spherical signals to discrete planar data through planar projection. However, this projection process inevitably entails distortion~\cite{ICLR:taco:s2cnn}.
CNNs learn local patterns by striding rectangular filters on the input data.
For this, CNNs assume that local patterns preserve their shapes even though their locations change by translations along the $x$ or $y$-axis.
This assumption is suitable for perspective images~\cite{alexnet, vgg, resnet}.
However, the assumption is violated in the case of spherical signals. 
If the spherical signals are transformed into perspective images via the planar projection, the projection process changes the shapes of local patterns according to their locations. 
In other words, there is no such a planar projection that can preserve the distance between two arbitrary points on the spherical signals.

The problem above occurs frequently in equirectangular projection (ERP). As shown in Fig.\ref{fig:PD_RD}, local patterns with the same shape can be transformed into significantly different shapes according to the latitudes in the course of ERP. The distortion happens not only in ERP but also in any type of planar projection. Because CNNs assume that a local pattern preserves its shape despite translation, the distortion entailed by the planar projection can considerably degrade the performance of CNNs.

The distortions can be divided into two types, projection and rotation distortions. 
(1) \textit{Projection distortion}. This distortion is defined as inconsistency of the shape of a local pattern. It is caused by the planar projection and salient in ERP. 
Fig.\ref{fig:PD_RD}(b) shows the projection distortion in ERP. 
(2) \textit{Rotation distortion}. This distortion is caused by the rotation of input data. If 3D rotation is applied to spherical signals, local patterns only change their locations on the sphere, as shown in Fig.\ref{fig:PD_RD}(c). However, a subsequent planar projection makes the local patterns with the same shape have different shapes, as shown in Fig.\ref{fig:PD_RD}(d). In other words, 3D rotation cannot be emulated as the 2D translation of local patterns~\cite{ICLR:taco:s2cnn}. Thus, CNNs that typically deal with 2D data lack equivariance toward 3D rotations.
The rotation distortion is a detrimental factor for representation learning. 
If the rotation distortion is perfectly eliminated, there exists $g^{\prime}$, which makes $f(g \cdot x) = g^{\prime} \cdot f(x)$ for the spherical signal $f$~\cite{SO3cnn} and the rotation operation $g$.
While the equality above cannot perfectly hold, we can measure the degree of equivalence using rotation equivariance errors~\cite{gcn_rotation, ICLR:taco:s2cnn}.

To overcome the aforementioned distortions of spherical signals projected on a plane, many researches have been introduced~\cite{khasanova2017graph, CVPR:lee:spherephd, ICLR:taco:s2cnn, SO3cnn, CVPR:su:ktn, gcn_rotation}.
In contrast to conventional methods, we first leverage the transformer architecture~\cite{ICLR:dosovitskiy:vit} to alleviate the effect of two distortions on 360\degree~images. 
Because we use the transformer architecture, we do not need a planar projection and can reduce the projection distortions.
In addition, we can reduce the rotation distortions for specific rotations, because the rotation equivariance can be achieved owing to the characteristics of the transformer architecture. 
To achieve this goal, we present a novel transformer for representation learning of 360\degree~images, which is called \textbf{\sphtr}.

Our contributions can be summarized as follows:

\begin{itemize}
\item To the best of our knowledge, it is the first to leverage the transformer architecture to perform representation learning for 360\degree~images. We demonstrate that the proposed transformer architecture is suitable for reducing the projection and rotation distortions. 

\item Our \sphtr~can be easily plugged into existing sampling methods. Using highly uniform sampling methods, our method can significantly reduce the projection distortion. For this, we define the uniformity measure to evaluate the uniformity of sampling methods.

\item  We leverage the permutation equivariance of \sphtr~and show that an element of the symmetry rotation group can be reduced to permutation of the input sequence. Thus, we argue that the rotation distortion can be effectively alleviated.
\end{itemize}

\begin{figure*}[t]
    \begin{minipage}[b]{0.24\linewidth}
		\centering
	\end{minipage}
	    \vspace{-3mm}
    \begin{minipage}[b]{\linewidth}
        \centering
        \includegraphics[width=\linewidth]{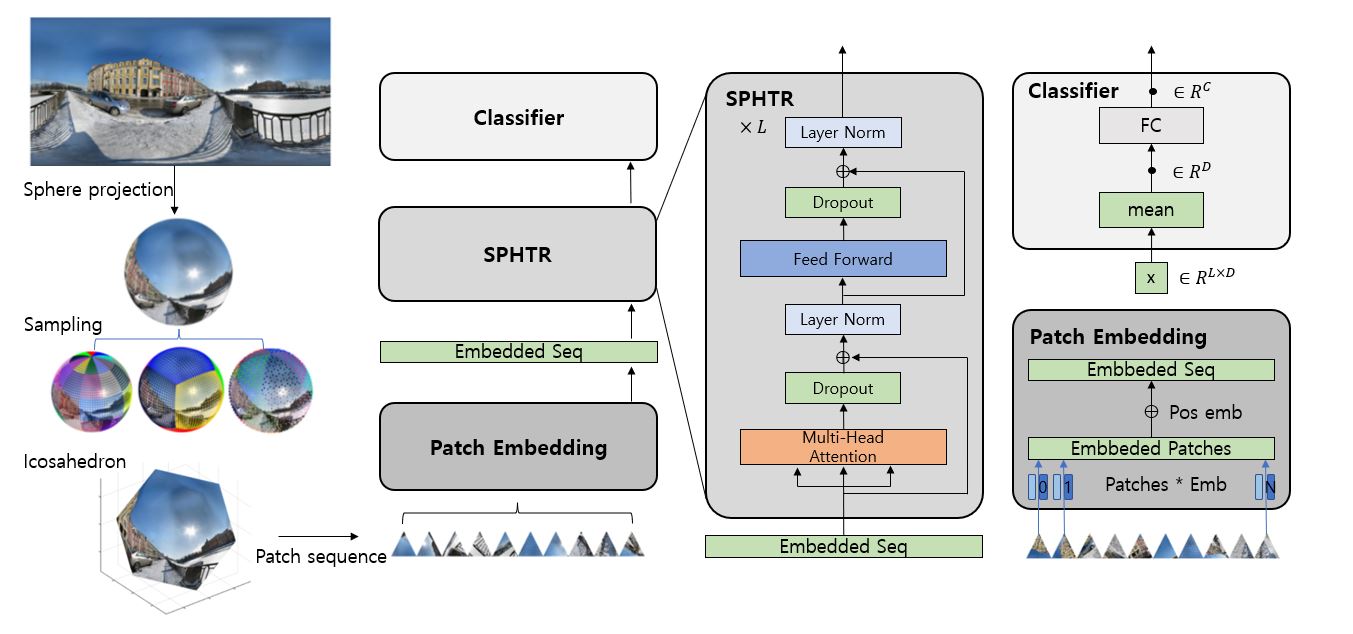}
    \end{minipage}
     \begin{minipage}[b]{0.24\linewidth}
		\centering
		(a)  
	\end{minipage}
	   \begin{minipage}[b]{0.24\linewidth}
		\centering
		(b)  
	\end{minipage}
	   \begin{minipage}[b]{0.24\linewidth}
		\centering
		(c)  
	\end{minipage}
    \begin{minipage}[b]{0.24\linewidth}
		\centering
		(d)  
	\end{minipage}
		    \vspace{-2mm}
    \caption{
    \textbf{The Overall procedure of the proposed method}. (a) Given a 360\degree~image visualized by ERP and spherical representation, we test three types of sampling methods. 
    Please note that the sampling method is not confined to three sampling methods but can be trivially extended without major modification of networks and the training algorithm. 
    (b) We sample points and group them into patches, in which the sequence of patches is an input of our \sphtr. 
    (c) \sphtr~is mainly based on ViT~\cite{ICLR:dosovitskiy:vit} to which the classification head in (d) is attached.
    }
    \label{fig:overall_pipeline}
    \vspace{-3mm}
\end{figure*}

\vspace*{-5mm}
\section{Related work}
\label{sec:related_work}

\noindent\textbf{Transformer.}
The transformer~\cite{ACL:devlin:bert, NIPS:vaswani:transformer} is designed to deal with sequential data for natural language processing tasks, in which it can learn the global relationship between vectors by using multi-head self-attention (MSA) and feed forward networks (FFN). Thus, the transformer exhibits better performance in terms of computation speed and accuracy compared to conventional RNNs~\cite{gated_rnn, lstm}. 

Visual transformer (ViT)~\cite{ICLR:dosovitskiy:vit} introduced a novel transformer encoder to process images rather than language data. 
By overcoming the limitations of conventional attention mechanisms~\cite{hu2019local}, ViT successfully replaced most portions of CNN structures with transformer blocks.
ViT showed superior performance to CNN counterparts with a smaller number of parameters for datasets that are larger than ImageNet~\cite{deng2009imagenet}. 
These good properties of the transformer inspired us to use the transformer architecture for processing 360\degree~images, as follows.
1) The transformer can learn the relationship between vectors along a sequential dimension. 
This property enables our \sphtr~based on the transformer to sample points from the surface of a sphere and group them into patches. Thus, \sphtr~can learn geometric characteristics between sampled points.
2) The transformer is suitable for learning visual representation from data, as demonstrated by ViT~\cite{ICLR:dosovitskiy:vit}. This property enables the proposed \sphtr~to learn visual representation of 360\degree~images.

Recently, the transformer architecture has been actively used for several computer vision tasks. For example, DETR~\cite{carion2020end} considered object detection problems as direct set prediction problems using the transformer architecture, in which hand-made anchors and the NMS stage were removed. 
Swin transformer~\cite{swin_transformer} modified vanilla transformers and made a general-purpose backbone network for processing visual data based on shifted window approaches.

\noindent\textbf{Representation learning of Spherical Signals.}
Representation learning for spherical signals has been steadily studied. 
Most researches focused on reducing distortions that can occur during the planar projections and learning features with the convolution operation.
S2CNN~\cite{ICLR:taco:s2cnn} defined CNNs on the $S^2$ and $SO(3)$ domains and proposed spherical CNNs. S2CNN was implemented using generalized Fourier transform and extracted features using spherical filters ($S^2, SO(3)$) on the input data, in which both expressive and rotation-equivariance were satisfied. S2CNN can be considered as an extension of 2D translation of conventional CNNs to spherical data with isomorphic filters to spheres.
SpherePHD~\cite{CVPR:lee:spherephd} applied CNNs to omni-directional images to resolve distortion problems. SpherePHD used regular polyhedrons to approximate 360\degree~images and projected 360\degree~images on the icosahedron that contains most faces among regular polyhedrons. This type of projection can reduce distortions compared to ERP and cube map projection. For this, novel convolution and pooling operations have been designed that suits for icosahedron.
SphereNet~\cite{ECCV:coors:spherenet} followed generic design of conventional CNNs with $3\times 3$ filters. However, SphereNet sampled pixels from highly uniform points on sphere surface compared to ERP and enhanced pooling layers to fit the convolution operation. 
KTN~\cite{CVPR:su:ktn} introduced a novel module that can compensate distortions occurred by the planar projection. KTN can use off-the-shelf CNNs trained on perspective images via model transfer, which enables to solve several recognition problems without re-training.
Methods based on graph convolution~\cite{khasanova2017graph, deepsphere, gcn_rotation} also achieved rotation equivariance.

In contrast to the aforementioned methods that use convolution to learn features, our method uses the transformer architecture for representation learning of spherical signals for the first time.
We leverage the advantages of the transformer architecture to reduce projection and rotation distortions. 
Our method is simple yet effective, in which a specialized library for implementation is not required.

\section{Proposed method}
\label{sec:proposed_method}
Section \ref{sec:sampling} presents sampling methods for 360\degree~images to build the input data of the proposed \sphtr. 
Section \ref{sec:uniformity} introduces a novel measure to evaluate the uniformity of the sampling method. 
Section \ref{sec:sphtr} describes the proposed transformer for 360\degree~image classification and shows the equivariance toward specific rotations. 
For this, we derive useful characteristics of regular polyhedrons that specific rotation can be reduced to a permutation of the input sequence. 

\subsection{Sampling Method} \label{sec:sampling}
\vspace{-2mm}
We introduce three sampling methods, which are ERP, cube, and icosahedron.
Each sampling method can be easily applied to \sphtr.
Please note that any sampling method that can be expressed by a sequence can be also applied.


\begin{figure}[t]
    \begin{minipage}[b]{0.24\linewidth}
		\centering
	\end{minipage}
      \vspace{-3mm}
    \begin{minipage}[b]{\linewidth}
	    \centering
        \includegraphics[width=\linewidth]{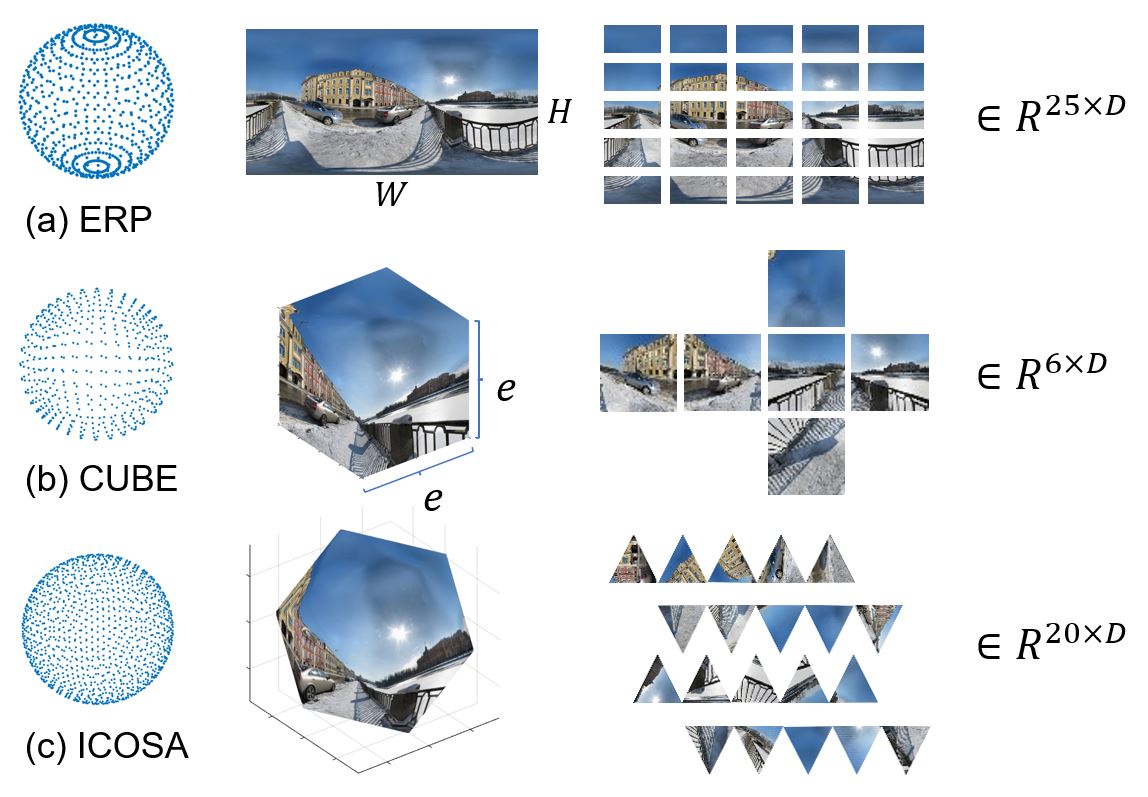}
    \end{minipage}
    \caption{\textbf{Three sampling methods}. (a) ERP, (b) Cube, (c) Icosahedron sampling methods.
    The leftmost figures represent the sampling points in the spheres, the middle figures show the projected images, and the rightmost figures illustrate the input patches for \sphtr~and their dimensions.}
    \label{fig:sampling_methods}
    \vspace{-6mm}
\end{figure}

\noindent{\bf Equirectangluar.}
The first sampling method is Equirectangluar (ERP).
ERP linearly maps the latitude vertically and the longitude horizontally with constant spacing. 
The latitude and longitude span $[0, \pi]$ and $[0, 2 \pi]$, respectively. 
Therefore, the projection has an aspect ratio of $1/2$. 
Fig.\ref{fig:sampling_methods}(a) shows ERP, of which shape is $(H, W)$. 
We sample points from each location, in which both $y$ and $x$ values are integers (pixel location) of ERP, and group them into patches. 
The size of each patch is $(P_w, P_h)$ and the corresponding input data has shape of  $\in \mathbb{R}^{N \times D}$, where $N = (H / P_h) \times (W / P_w)$ and $D = (H \times W)/N$.

\noindent{\bf Cube.}
The cube sampling method is based on the cube map projection. The cube map projection approximates a sphere onto a circumscribing cube. We sample pixel points from pixel locations of the cube map, in which the dimension of each face is $e \times e$. Then, as shown in Fig.\ref{fig:sampling_methods}(b), we flatten each patch to obtain the input sequence with the size of $\mathbb{R}^{N \times D}$, where $N = 6$ and $D = e^{2}$.

\begin{figure}[t]
    \begin{minipage}[b]{0.24\linewidth}
		\centering
	\end{minipage}
      \vspace{-3mm}
    \begin{minipage}[b]{\linewidth}
		\centering
		    \includegraphics[width=0.9\linewidth]{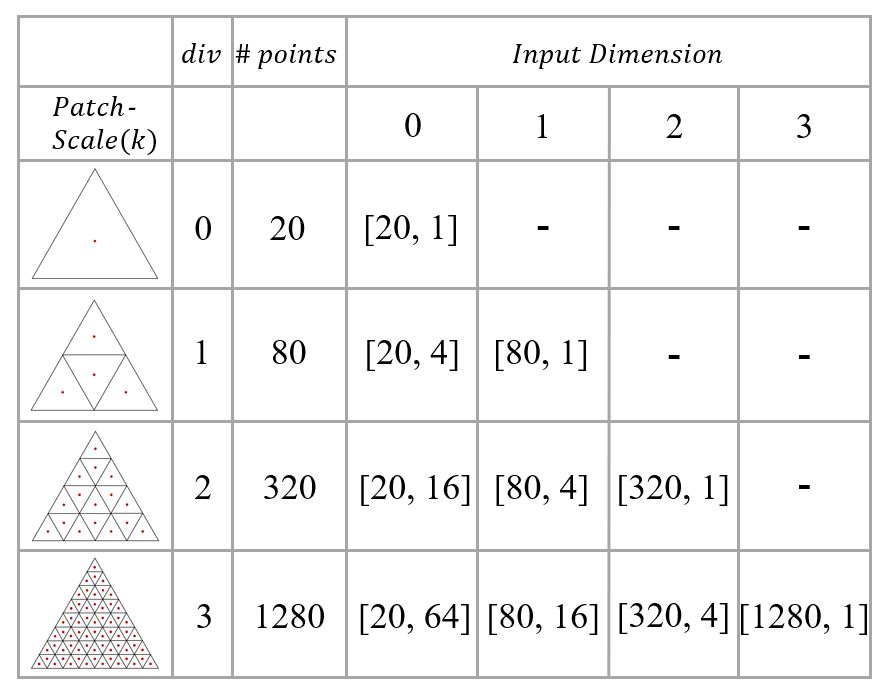}
	\end{minipage}
    \caption{\textbf{Icosahedral subsampling}. The icosahedron can be divided into $20$ patches on the sphere. We can make $4$ small triangles from one large triangle by joining the midpoints of each side and extend to their centers. 
    We can also use the patch scale to reshape the points into various patches, where div means the division level.}
    \vspace{-3mm}
    \label{fig:icosahedral_subsampling}
\end{figure}

\noindent{\bf Icosahedron.}
Icosahedron has the most number of faces (20 faces) among regular polyhedrons.
A sphere can be projected onto an inscribing icosahedron. 
Each face of the icosahedron can be hierarchically and iteratively divided into $4$ congruent equilateral triangles by connecting midpoints of each side, as shown in Fig.\ref{fig:icosahedral_subsampling}.
The leftmost column of Fig.\ref{fig:icosahedral_subsampling} describes the shape and point of the patch according to the division levels (\ie the number of divisions).
We sample points from the center of each triangle and group them according to the faces of the icosahedron.
Thus, if we apply the subdivision for $div$ times, we obtain the input data $\mathbb{R}^{N \times D}$, where $N=20$ and $D=4^{div}$.
According to the division level ($div$), $20 \times 4^{div}$ points can be sampled from the icosahedron.
To make the input for \sphtr, we reshape the length of the sequence according to the patch scale $k$. 
Then, the input dimension is $[20 \times 4^k, 4 ^ {div - k}]$.
Fig.\ref{fig:icosahedral_subsampling} shows the input dimension according to $k$.


\subsection{Uniformity Measure} \label{sec:uniformity}
We defined the projection distortion as the inconsistency of local patterns according to their locations. 
Then, a major cause of the distortion and inconsistency is an uneven distribution of pixels (or points) on the sphere. 
For example, ERP has a dense distribution of pixels on two polar regions, but the distribution becomes sparser as it approaches the equator. 
If an object approaches to two polar regions, the object becomes enlarged in ERP. 
Thus, if points or pixels are distributed uniformly, the projection distortion can be alleviated. 
In this context, we consider the uniformity value as a proxy of the projection distortion.
If we use a sampling method with high uniformity, we can mitigate the projection distortion.

\begin{figure}[t]
    \begin{minipage}[b]{0.24\linewidth}
		\centering
	\end{minipage}
      \vspace{-3mm}
    \begin{minipage}[b]{\linewidth}
		\centering
        \includegraphics[width=\linewidth]{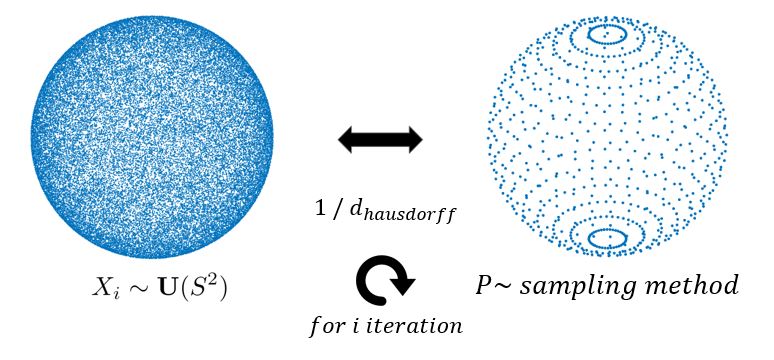}
     \end{minipage}
    \caption{\textbf{Computation of the uniformity measure}. 
    $\{X_i\}$ denotes the points, which are sampled from the uniform distribution defined on the 3D-sphere, $\mathbf{U}(S^2)$. 
    $P$ denotes the points obtained by a sampling method, which is to be measured. 
    We define the uniformity as the inverse of the distance between $\{X_i\}$ and $P$. The uniformity can be computed by iteratively sampling $X_i$ from the uniform distribution. The convergence test of the uniformity is conducted in Section \ref{sec:exp_Uniformity}.}
    \vspace{-4mm}
    \label{fig:uniformity_measure}
\end{figure}

\begin{definition}[Uniformity]
The Uniformity is defined as
\begin{equation}
\label{uniformity}
    Unif(P) = \frac{1}{n}\sum_{i=1}^{n}(d_{hd}(X_i, P)^{-1})\text{, for } X_i \sim \mathbf{U}(S^2),
\end{equation}
where $d_{hd}$ is the Hausdorff distance between two sets and $\mathbf{U}$ is a uniform distribution on the 3D-sphere~\cite{muller1959note}. The uniformity measure is an expected value of the inverse of the distance between the set of points sampled from the ideal distribution (\ie $\{X_i\}_{i=1}^n$) and the set of points from the sampling method (\ie $P$), as shown in Fig.\ref{fig:uniformity_measure}. 
\end{definition}

The Hausdorff distance $d_{hd}$~\cite{attouch1991topology} is the largest distance between a point in one set to the closest point in another set, which is formulated as follows:
\begin{equation}
\label{hausdorff distance}
    d_{hd}(X, Y) = \max\left\{\sup_{x \in X} d(x, Y),\sup_{y \in Y} d(X, y)\right\}, 
\end{equation}
where $d$ denotes the Euclidean distance and $d(a,B)=\inf_{b\in B}d(a,b)$.
Because $d_{hd}(P, X)$ is the largest value of the minimum distances between all points of each set, it can be seen as a kind of supremum (the upper limit) of the distance, which elements of two sets can have. 
Thus, as this distance becomes small, the uniformity measure has a large value. 
Large uniformity values indicate highly uniform distributions of the points set $P$.


\subsection{Spherical Transformer} \label{sec:sphtr}
We propose a novel spherical transformer (\sphtr), that is a transformer for representation learning of spherical signals. 
\sphtr~has several advantages when learning spherical signals. 
First, \sphtr~can easily adopt a sampling strategy to mitigate distortions.
Second, \sphtr~can achieve the rotation equivariance to a finite set of rotations by extending the characteristics of the transformer's permutation equivariance. 
In this section, we first describe the structure of \sphtr~and show its rotation equivariant property.

\noindent{\bf Network Architecture.}
Transformer~\cite{NIPS:vaswani:transformer} requires the sequence vector $x \in \mathbb{R} ^ {N \times D}$ as an input.
$x$ can be divided into $N$ patches, $x = \{x_p^1 ... x_p^N\}$, where $N$ is the length of sequence and $D$ is the dimension of the input.     
Fig.\ref{fig:overall_pipeline} shows the overall structure of \sphtr, where MSA and FFN are explained in detail in the Supplementary Material and CLF is a classification Head.

\noindent$\bullet$ \textit{Patch Embedding in Fig.\ref{fig:overall_pipeline}(b) and (d)}: the input sequence is embedded to the latent vector and positional embedding is conducted, which results in the embedded sequence.
\begin{equation}
\label{eqn:embedding}
    s_0=[x^1_p \textbf{E} ;...; x^{N}_p \textbf{E}] + \textbf{E}_{pos}, 
\end{equation}
where $\textbf{E}$ $\in \mathbb{R}^{D \times D_m}$ is the linear mapping matrix, $E_{pos}$ $\in \mathbb{R}^{N \times D_m}$ is learnable parameters for the positional embedding, and $D_m$ is the model dimension.
We do not use the [cls] token in Bert~\cite{ACL:devlin:bert}, which assumes that a specific token contains information on the entire sequence. 
Instead, we use the representation from all patches to keep the orientation information in every single patch.

\noindent$\bullet$ \textit{Multi-head Self-attention (MSA) in Fig.\ref{fig:overall_pipeline}(c)}: given the embedded sequence, multi-head self-attention~\cite{ICLR:dosovitskiy:vit, NIPS:vaswani:transformer} is conducted.
\begin{equation}
\label{eqn:msa}
    s_l^{'}=MSA(LN(dropout(s_{l-1}))) + s_{l-1}. 
\end{equation}

\noindent$\bullet$ \textit{Feed Forward Network Layer (FFN) in Fig.\ref{fig:overall_pipeline}(c)}: subsequently, the feed forward layer and the Post-LN transformer layer~\cite{xiong2020layer} are used to make features, where the LN is located behind the MSA.
\begin{equation}
\label{eqn:ffn}
    s_l=FFN(LN(dropout(s_l^{'}))) + s_l^{'}.
\end{equation}

\noindent$\bullet$ \textit{Classification Layer (CLF) in Fig.\ref{fig:overall_pipeline}(d)}: 
As the last step, classification layer is to make the logit for the output.
For this, we use one linear layer. Before passing the layer, the average is obtained in the direction of the length.
Our method uses the average for classification, whereas other transformers use [cls] token features.
\begin{equation}
\label{eqn:clf}
    y =CLF[LN(s_{l=L})].  
\end{equation}

\begin{figure}[t]
    \begin{minipage}[b]{0.24\linewidth}
		\centering
	\end{minipage}
      \vspace{-3mm}
    \begin{minipage}[b]{\linewidth}
		\centering
        \includegraphics[width=\linewidth]{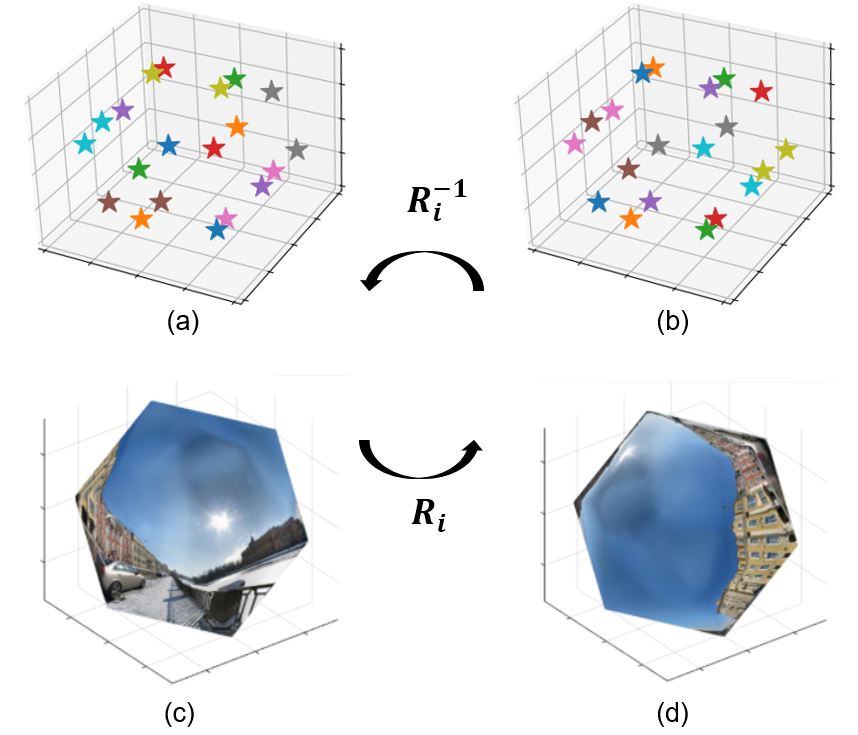}
    \end{minipage}
    \caption{\textbf{Elements of symmetry rotation groups of icosahedron are isometries}.
     (a) $20$ center points of the face of the icosahedron.
     (b) $R_i$, an element of the symmetry rotation group, is multiplied to the points in (a), in which the positions of the points preserve despite the rotation $R_i$ and only the order is changed.In other word, Isometry preserves the distance between any two points after the rotation. Isometry of regular polyhedron results in permutation of faces.
     (d) $20$ patches after the rotation of those in (c).
     }
    \label{fig:icosahedral isometry}
    \vspace{-2mm}
\end{figure}

\noindent{\bf Rotation Equivariance of SPHTR.}
We demonstrate that \sphtr~is equivariant to a finite set of rotations. In particular, the finite set is a symmetry rotation group that is associated to the regular polyhedron based sampling methods.

\begin{definition}[equivariance~\cite{SO3cnn}]
\label{def:equivarinat}
Let $f : X \rightarrow Y$ be a function and $G$ be a group. 
We assume that $G$ acts on $X$ and $Y$.
$L_g$ denotes the group operator that takes a function $f$ and produces the translated function $L_gf$.
The function $f$ is equivariant to transformations $g \in G$ 
if $[f(L_gx)] = [L_g(f(x))] \quad \forall x \in X, \forall g \in G$.
\end{definition}
According to Yun et al.~\cite{yun2019transformers}, the transformer is equivariant to permutation of the input sequence. With Definition~\ref{def:equivarinat}, we can derive the following equality. 
\begin{equation}
\label{eq:permutation equivariant}
    [f(L_\pi x)] (p) = [L_\pi(f(x))] (p),
\end{equation} 
where $L_\pi$ is a permutation operator and $p$ is a point on the sphere.
A regular polyhedron has a symmetry rotation group that can maintain its position and shape even after the rotation.
For example, icosahedron has a symmetry rotation group called icosahedral rotation group $R_i$ of order $60$, which is
a subgroup of SO(3)~\cite{gcn_rotation}.
Fig.\ref{fig:icosahedral isometry} (a) and (b) show that the positions of the points preserve despite the rotation $R_i$ and only the order is changed.
In other words, there is a corresponding permutation operator $\pi_i$ for every rotation $R_{i}$ and a permutation is applied to faces of the icosahedron (\ie $L_{R_i}\leftrightarrow L_{\pi_i}~for~R_i, \pi_i \in \Pi $).
Thus, \sphtr~has rotation equivariant properties for a specific rotation. 
\begin{equation}
\label{eq:permutation equivariant}
    [f(L_{R_i} x)] (p) = [L_{R_i}(f(x))] (p).
\end{equation}

\begin{figure}[t]
    \begin{minipage}[b]{0.24\linewidth}
		\centering
	\end{minipage}
      \vspace{-5mm}
    \begin{minipage}[b]{\linewidth}
		\centering
        \includegraphics[width=\linewidth]{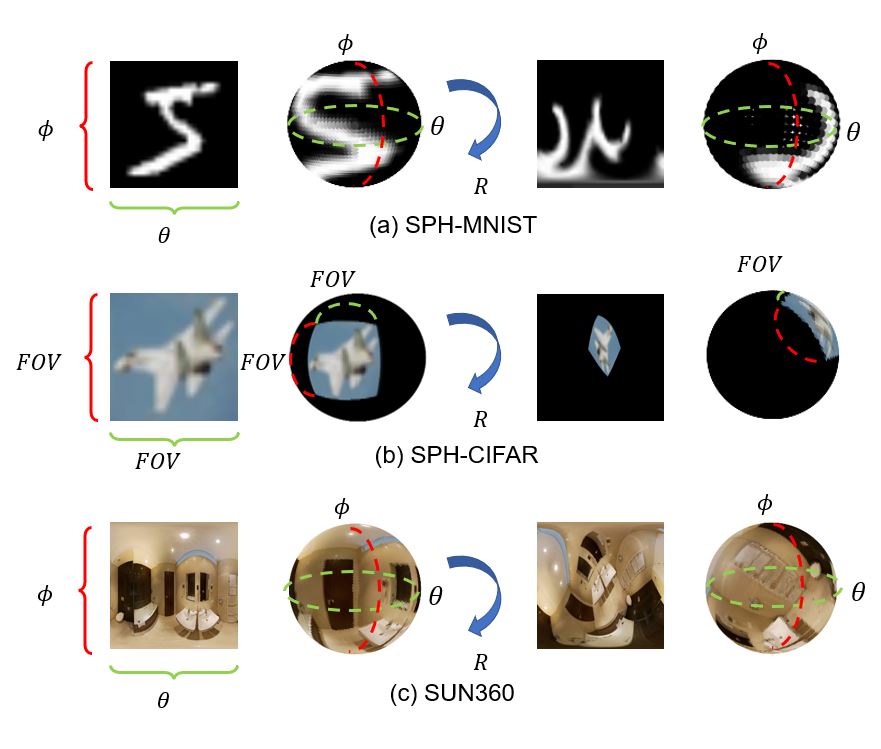}
    \end{minipage}
    \caption{\textbf{Datasets}. (a) SPH-MNIST (Left: MNIST and Right: rotated MNIST). The MNIST image is projected to the entire sphere. (b) SPH-CIFAR. The rotated SPH-CIFAR image is mapped to the sphere. At this time, we project the image into a part of the sphere considering the field of view (FOV). (c) SUN360 dataset consisting of images with the locations and the corresponding labels. We project these images into the entire sphere.
    }
    \label{fig:datasets}
\end{figure}

\begin{table}[t] 
\caption{\textbf{Network configuration of \sphtr} for each experimental dataset, in which we considered the complexity of the dataset and the resolution of the input.}
\label{tab:model_config}
\vspace{-3mm}
\setlength{\tabcolsep}{3pt}
\centering
{\footnotesize
\begin{tabular}{cccccc} 
\hline 
dataset     & epochs    & model dim & $\#$ Layers & \# Heads & \# Params \\ \thickhline 
SPH-MNIST   & 200       & 24        & 8           & 8        & 60k       \\ \hline 
SPH-CIFAR   & 100       & 24        & 4           & 8        & 60k       \\ \hline 
SUN360      & 100       & 192       & 8           & 8        & 251k      \\ \hline
\end{tabular}
}
\end{table}

\begin{table*}[t]
\caption{\textbf{Evaluation  of uniformity and sampling method}. We trained \sphtr~with different sampling methods and a baseline CNN. Icosahedron reports the highest uniformity for both datasets. Please note that the ranks of accuracy and uniformity coincide.}
\label{tab:uniformity}
\centering
\vspace{-3mm}
\setlength{\tabcolsep}{8pt}
\begin{tabular}{|c|c|c|c|c|c|c|c|c|}

\hline
Datasets                  & Sampling & Model & Params & \# Point & Uniformity & Accuracy & Unif. rank & Acc. rank \\ \thickhline
\multirow{4}{*}{SPH-MNIST}& ERP     & CNN   & 61.2k    & 1250     & -         & 0.6812   & -     & -                                 \\ \cline{2-9} 
                         & ERP      & SPHTR & 59.9k      & 1250     & 8.275    & 0.2441   & 3     & 3        \\ \cline{2-9} 
                         & CUBE     & SPHTR & 63.6k      & 1350     & 10.114   & 0.8799   & 2     & 2        \\ \cline{2-9} 
                         & ICOSA    & SPHTR & 60.0k      & 1280     &\bf{10.6338}&\bf{0.9043}& 1  & 1        \\ \thickhline
\multirow{4}{*}{SPH-CIFAR} & ERP      & CNN   & 86.0k  & 5000     & -          & 0.2956  & -     & -     \\ \cline{2-9} 
                         & ERP      & SPHTR & 44.2k      & 5000     & 18.0280    & 0.1319  & 3     & 3        \\ \cline{2-9} 
                         & CUBE     & SPHTR & 85.8k      & 5046     & 20.2055    & 0.3142  & 2     & 2        \\ \cline{2-9} 
                         & ICOSA    & SPHTR & 48.1k      & 5120     &\bf{21.2247}& \bf{0.3784} & 1  & 1        \\ \hline
\end{tabular}
\vspace{-3mm}
\end{table*}

\section{Experiments}
\label{sec:experiments}
Section \ref{sec:exp_settings} describes the experimental settings and implementation details.
Section \ref{sec:exp_Uniformity} contains the evaluation and convergence analysis of Uniformity. Section \ref{sec:exp_Equivariance} shows the computation of rotation equivariance error. Section \ref{sec:ablation} conducted the ablation study for the proposed method. Section \ref{sec:comparison} compares our method with other algorithms.

\subsection{Experimental Settings} \label{sec:exp_settings}

\noindent{\bf Dataset.}
We constructed synthetic spherical images using the MNIST and CIFAR10 datasets~\cite{ICLR:taco:s2cnn, CVPR:lee:spherephd, gcn_rotation}, as shown in Fig.\ref{fig:datasets}, which are called \smnist~and \scifar, respectively.
\smnist~was mapped onto the whole sphere and \scifar~was projected with the $65.5$\degree~of horizontal and vertical FOV. With these synthetic 360\degree~images, we also used the natural 360\degree~image dataset called SUN360. SUN360~\cite{sun360_dataset} is composed of around $60,000$ full spherical images with the resolution of $512\times 1024$ and place information labels. We carefully chose $10$ visually distinguishable classes. The images have been divided into $1704$ train and $200$ test images. All training and testing images are randomly rotated.
For the number of icosahedron samples for each dataset, 1280, 5120, and 20480 were selected considering the resolution of each dataset.

\noindent{\bf Implementation Details.}
Table~\ref{tab:model_config} shows our model configuration for each experiment.
Adam optimizer was used for all experiments. 
The initial learning rate was set to $10^{-3}$ for \smnist~and \scifar, and $10^{-4}$ for SUN360. The batch size was set to $128$ and a cosine annealing scheduler was used~\cite{ACL:devlin:bert}. All the code was written in Pytorch 1.8.1. 

\noindent{\bf Evaluation.}
We solved 360\degree~image classification problems using various datasets, in which each dataset has $10$ classes and includes different resolutions. 
The accuracy in Tables \ref{tab:uniformity}, \ref{tab:ablation_patch_scale}, and \ref{tab:comparison} denotes the classification top-1 accuracy.

\begin{figure}[t]
    \begin{minipage}[b]{0.24\linewidth}
		\centering
	\end{minipage}
      \vspace{-3mm}
    \begin{minipage}[b]{\linewidth}
		\centering
         \includegraphics[width=\linewidth]{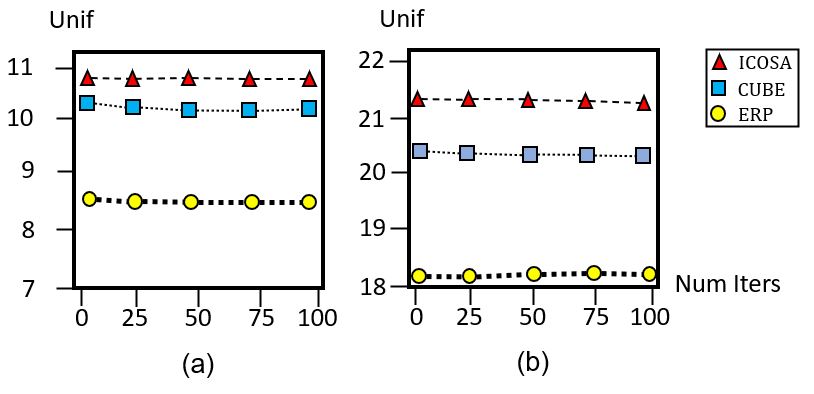}
    \end{minipage}
    \caption{\textbf{Uniformity measure of different sampling methods}. $x$-axis is the number of iterations and $y$ axis is uniformity measure. As shown in the figure, uniformity measure converges to a certain number as the iteration goes on, which supports the validity of our uniformity measure.}
    \label{fig:unif_convergence}
    \vspace{-5mm}
\end{figure}

\subsection{Experiments on Uniformity} \label{sec:exp_Uniformity}
\noindent{\bf Evaluation of Uniformity and Sampling Method.}
The uniformity measure acts as a proxy for alleviation of the projection distortion. We computed the uniformity measure and trained \sphtr~with three sampling methods. Table~\ref{tab:uniformity} reports information on the uniformity measure and performance on image classification tasks. The number of points and network parameters were adjusted to have similar values. Icosahedron reported the highest uniformity followed by cube and ERP for both datasets. We can confirm that the uniformity measure coincides with the performance for \smnist~and \scifar~(please compare Unif. rank with Acc. rank columns), which supports the validity of the uniformity measure as the proxy of the projection method. 

One can notice that the uniformity measure in the \scifar~setting is higher than that of the \smnist~setting. This is due to the characteristics of the Hausdorff distance that shows the decreasing tendency as the numbers of points for two sets increase. However, in the setting where the numbers of points are similar, we can use the uniformity measure as an indicator of relative discrepancy to others. Fig.\ref{fig:unif_convergence} shows that there is a consistent gap between sampling methods when the number of points is fixed.

\begin{figure}[t]
      \begin{minipage}[b]{\linewidth}
		\centering
        \includegraphics[width=\linewidth]{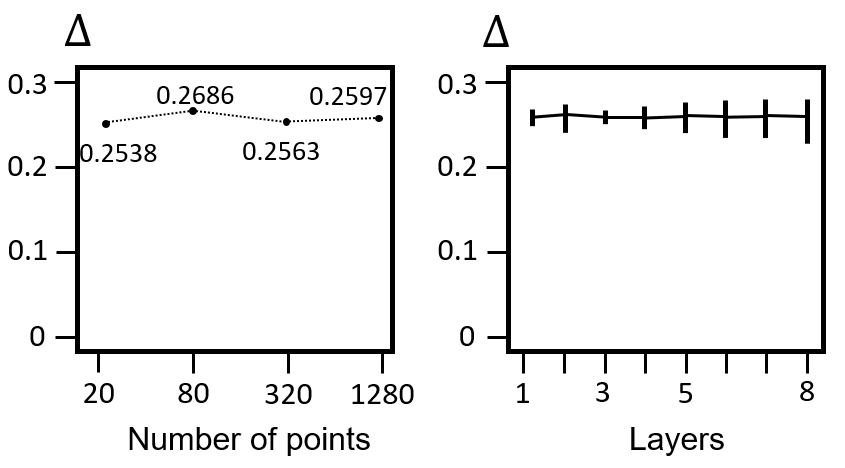}
    	\end{minipage}
      \begin{minipage}[b]{0.49\linewidth}
		\centering
		(a)  
	\end{minipage}
	\begin{minipage}[b]{0.49\linewidth}
		\centering
		(b)  
	\end{minipage}
	\vspace{-3mm}
    \caption{\textbf{Equivariance error}. (a) equivariance error according to the number of points for one layer, (b) equivariance error according to the number of layers. $\Delta$ is the equivariance error.
    }
    \label{fig:equivariance_error}
    \vspace{-5mm}
\end{figure}

\noindent{\bf Convergence of Uniformity.}
We computed the uniformity by running and taking the average value of the iterations according to Definition \ref{uniformity}. If the uniformity measure does not converge to a specific value under as the number iterations increases, the uniformity lacks its validity. Therefore, we reported the uniformity with varying numbers of iterations. We run the uniformity until $100$ iterations and report the value with the step size of $25$. Fig.\ref{fig:unif_convergence} shows that the uniformity converges to a specific value.

\subsection{Equivariance Error} \label{sec:exp_Equivariance}
We reported the rotation equivariance error~\cite{ICLR:taco:s2cnn} for the rotations of symmetry rotation group to validate our argument. 
The equivariance error is computed as follows.
\begin{equation}
\label{equ:equivariance error}
    \Delta = \frac{1}{n}\sum_{i=1}^{n}std(L_{R_{i}}\Phi(x_i) - \Phi(L_{R_{i}} x_i)) / std(\Phi(x_i)).
\end{equation}
$\Phi$ is stacked with encoder layers, in which the model dimension is $16$ and there are $8$ heads. 
$R_{i}$ is a rotation that belongs to a symmetry rotation group of icosahedron, which has order of $60$.
$L_{R_{i}}$ denotes the rotation operator, 
In our experiments, $n = 1000$ and $x_i$ was sampled from the SPH MNIST dataset.  
The sampling method was chosen as the icosahedron, because it shows the best performance and belongs to a regular polyhedron. We computed the rotation equivariance error with various division levels and layers. 
Fig.\ref{fig:equivariance_error} (a) and (b) show the equivariance errors according to the number of points (resolution) and the number of layers, respectively.
Thus, this result implies that our \sphtr~has rotation equivariance regardless of the number of points and layers.


\begin{figure}[t]
    \includegraphics[width=\linewidth]{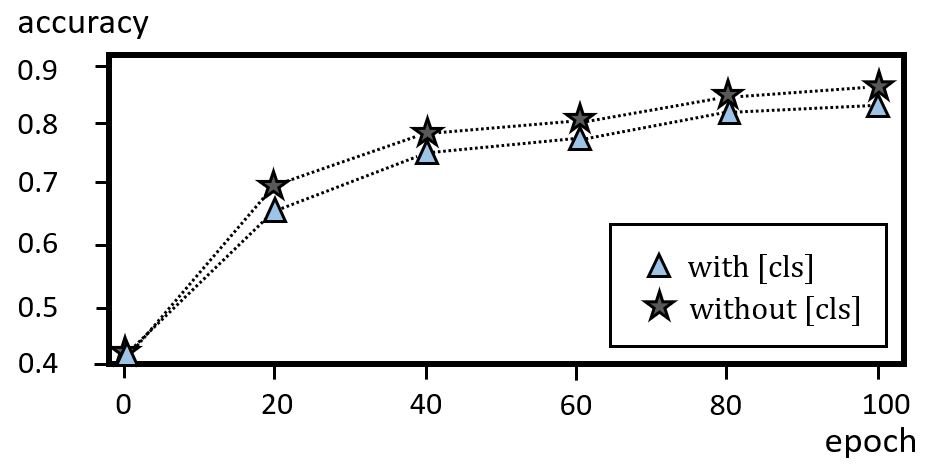}
    \vspace{-7mm}
    \caption{\textbf{Ablation 1: with/without [cls] tokens}. This ablation study is related to [cls] tokens. Orange means without [cls] tokens and blue means with [cls] tokens. As we assume to keep information about each direction, the continuous learning accuracy was higher when class tokens were subtracted.}
    \vspace{-4mm}
    \label{fig:cls_token}
\end{figure}

\subsection{Ablation Study} \label{sec:ablation}
We conducted ablation study on each component of the proposed \sphtr~to demonstrate the efficiency of \sphtr.

\noindent{\bf Cls Token of SPHTR}. 
Our \sphtr~does not use [cls] token. 
We have two main reasons for this.
First, if [cls] token is applied, then the rotation equivariance error is difficult to be measured according to \eqref{equ:equivariance error}. 
The second reason is to keep the orientation information present in every single patch for rotation.
Fig.\ref{fig:cls_token} shows the difference in performance with/without the [cls] token in SPH-MNIST dataset classification with Icosahedron sampling methods. 
The network without the [cls] token is not used shows better performance.

\noindent{\bf Patch Scale $k$}.
As shown in Fig.\ref{fig:icosahedral_subsampling}, using the patch scale of the Icosahedron sampling method, it was possible to reshape the same number of sampled points with input dimensions of various shapes.
In this ablation study, we experiment \sphtr~according to various patch scale $k$. 
Table \ref{tab:ablation_patch_scale} reports performance of \sphtr~with various patch scale $k$, in which we can observe a positive correlation between patch scale $k$ and accuracy for \scifar.

\begin{table}[t] 
\caption{\textbf{Ablation 2: different patch scale $k$}. This experiment shows the effect of different values about patch scale $k$ for SPH-cifar dataset and icosahedron sampling method.}
\label{tab:ablation_patch_scale}
\vspace{-1mm}
\centering
\begin{tabular}{|c|c|c|c|c|}
\hline
Dataset                    & k & params & Input shape    & Accuracy      \\ \thickhline
\multirow{4}{*}{SPH-CIFAR} & 0 & 48.1k  & {[}20, 768{]}  & 0.3784        \\ \cline{2-5} 
                           & 1 & 35.7k  & {[}80, 192{]}  & 0.4565        \\ \cline{2-5} 
                           & 2 & 38.0k  & {[}320, 48{]}  & 0.5275        \\ \cline{2-5} 
                           & 3 & 60.2k  & {[}1280, 12{]} & \bf{0.5821}   \\ \hline
\end{tabular}
\vspace{-3mm}
\end{table}

\subsection{Comparison} \label{sec:comparison}
We compared our method with S2CNN \cite{ICLR:taco:s2cnn}, SpherePHD \cite{CVPR:lee:spherephd}, and SGCN \cite{gcn_rotation} using three datasets. We considered \smnist~and \scifar~as a low resolution dataset and SUN360 as a high resolution dataset. We trained $200$, $100$, and $100$ epochs for \smnist, \scifar,~and SUN360, respectively. For this experiment, we used publicly available codes of original authors.
We followed the original paper and code from the authors for other settings such as optimizer or scheduler.

\noindent{\bf Comparison Using Low Resolution Datasets}. 
Because \smnist~and \scifar~has low resolution, we balanced all the networks to have a similar number of parameters (around 60k to 70k). Table~\ref{tab:comparison} shows the result for both \smnist~and \scifar. SGCN and SpherePHD performed on par on \smnist. Please note that our network can have space for improved performance with extended duration, thus we report additional accuracy in (ours(*)). With the extended duration, our method performed better than others. 
Due to the nature of the transformer~\cite{ICLR:dosovitskiy:vit} the inductive bias is insufficient, so it takes longer to learn its geometry than CNNs.
For the SPH-CIFAR dataset, our method, SGCN, and SpherePHD exhibited similar performance. 
However, S2CNN$^{\dagger}$ showed 10\% of accuracy, which is similar to random prediction. It should be noted that we used the network structure and code of the original author.


\begin{table}[]
\caption{\textbf{Comparison between our method and other state-of-the-art algorithms}. We report the additional accuracy using the extended training duration for the SPH-MNIST dataset in ours(*). 
The best results were written in boldface.
}
\label{tab:comparison}
\centering
\vspace{-3mm}
\begin{tabular}{|c|c|c|c|}
\hline
Dataset                    & Methods   & Params & Accuracy         \\ \thickhline 
\multirow{5}{*}{SPH-MNIST} & SpherePHD~\cite{CVPR:lee:spherephd} & 57k    & 0.9408           \\ \cline{2-4} 
                           & S2CNN~\cite{ICLR:taco:s2cnn}        & 58k    & 0.9303           \\ \cline{2-4} 
                           & SGCN~\cite{gcn_rotation}            & 60k    & 0.9442           \\ \cline{2-4} 
                           & Ours      & 60k    & 0.9043           \\ \cline{2-4} 
                           & Ours(*)   & 60k    & \textbf{0.9509}  \\ \thickhline
\multirow{4}{*}{SPH-CIFAR} & SpherePHD~\cite{CVPR:lee:spherephd} & 57k    & 0.592            \\ \cline{2-4} 
                           & S2CNN$^{\dagger}$~\cite{ICLR:taco:s2cnn}        & 59k    & 0.1              \\ \cline{2-4} 
                           & SGCN~\cite{gcn_rotation}            & 58k    & \textbf{0.6072}  \\ \cline{2-4} 
                           & Ours                                & 60k    & 0.5821           \\ \thickhline
\multirow{4}{*}{SUN360}    & SpherePHD~\cite{CVPR:lee:spherephd} & 280k   & 0.580            \\ \cline{2-4} 
                           & S2CNN~\cite{ICLR:taco:s2cnn}        & 544k   & 0.655            \\ \cline{2-4} 
                           & SGCN~\cite{gcn_rotation}            & 433k   & 0.675            \\ \cline{2-4} 
                           & Ours      & 251k   & \textbf{0.7}     \\ \hline
\end{tabular}
\end{table}

\noindent{\bf Comparison Using High Resolution Dataset}. 
SUN360 is composed of natural full spherical images with high resolution and includes more complicated information compared to SPH-MNIST and SPH-CIFAR, as shown in Fig.\ref{fig:datasets}. Therefore, we scaled up all the networks to have more parameters for experiments on the SUN360 dataset. The results are shown in Table \ref{tab:comparison}. Our method shows the best performance among other algorithms followed by SGCN, S2CNN, and SpherePHD, while the proposed method requires the smallest number of parameters. SGCN, S2CNN, and SpherePHD were scaled up to the limitation, where particular graphic card settings (around 10 gigabytes) were required. We analyzed the memory efficiency of our method against other algorithms in the Supplementary Material.


\section{Limitation and Future Work}
The transformer outperforms its CNN counterparts when the dataset is large enough. However, we had to use small datasets to train \sphtr, due to the lack of large 360\degree~image datasets for image classification tasks. In addition, the rotation equivariance is confined to the rotation symmetry group of regular polyhedrons. By considering these limitations, we can present two future works. First, a gigantic dataset consisting of millions of natural 360\degree~images can be used to improve the performance of the proposed \sphtr~and its variants. Second, rotation symmetry group of regular polyhedrons and permutation equivariance of the transformer architecture can be more thoroughly employed based on the field of geometric deep learning and group theory.  

\section{Conclusion}
Applying CNNs for 360\degree~images results in sub-optimal performance due to distortions entailed by a planar projection. The distortion gets deteriorated when a rotation is applied to the 360\degree~images. Therefore, many researches based on convolution focus on reducing the distortions to learn representation.
We leverage the transformer architecture in order to perform image classification on the 360\degree~images. Using transformer architecture for the 360\degree~images have two main advantages. First, we do not need planar projections by sampling pixels from the sphere surface. Second, for sampling methods based on regular polyhedrons, specific rotations can be reduced to the permutation of faces, thus achieving low rotation equivariance errors.
Experimental results demonstrate that our method is competitive to other state-of-the-art algorithms on the SPH-MNIST, SPH-CIFAR, and SUN360 datasets.

{\small
\bibliographystyle{ieee_fullname}
\bibliography{egbib}
}
\end{CJK}
\end{document}